# Artificial Rigidities vs. Biological Noise: A Comparative Analysis of Multisensory Integration in AV-HuBERT and Human Observers

Francisco Portillo López[a]

[A] fportillolo@alumni.unav.es

*Universidad de Navarra, Spain*

**ABSTRACT**

This study evaluates AV-HuBERT's perceptual bio-fidelity by benchmarking its response to incongruent audiovisual stimuli (McGurk effect) against human observers ($N=44$). Results reveal a striking quantitative isomorphism: AI and humans exhibited nearly identical auditory dominance rates (32.0% vs. 31.8%), suggesting the model captures biological thresholds for auditory resistance. However, AV-HuBERT showed a deterministic bias toward phonetic fusion (68.0%), significantly exceeding human rates (47.7%). While humans displayed perceptual stochasticity and diverse error profiles, the model remained strictly categorical. Findings suggest that current self-supervised architectures mimic multisensory outcomes but lack the neural variability inherent to human speech perception.

## 1. Introduction

The human speech perception is inherently multisensory, integrating auditory and visual information to facilitate accurate recognition in ambiguous environments. Visual cues, particularly lip movements, have been recognized as critical for speech perception, enhancing intelligibility under noisy conditions (Sumby & Pollack, 1954; Massaro, 1987). A striking demonstration of audiovisual integration is the McGurk effect, in which incongruent auditory and visual cues produce a perception distinct from either input separately. For instance, an auditory /ba/ stimulus combined with a visual articulation of /ga/ typically generates the perception of /da/ (McGurk & MacDonald, 1976). This phenomenon highlights the obligatory nature of multisensory integration in speech perception and has been replicated across different languages and experimental paradigms (Rosenblum, 2008; Schwartz, Robert-Ribes & Escudier, 1998).

The McGurk effect has also informed the development of computational models of speech perception. Probabilistic frameworks, especially Bayesian models, suggest that the brain dynamically weights sensory inputs according to their reliability to resolve audiovisual conflicts (Ernst & Banks, 2002; Körding et al., 2007; Alais & Burr, 2004). More recently, causal inference models have been proposed to explain variability in McGurk effect susceptibility across individuals and contexts, postulating that the perceptual system first infers whether auditory and visual signals originate from the same source before integrating them (Magnotti & Beauchamp, 2017). These models provide a theoretical basis for quantifying perceptual outcomes under controlled experimental conditions.

In parallel, recent advances in artificial intelligence have led to audiovisual speech recognition models that combine visual cues with acoustic information to improve performance, particularly in noisy environments. Multimodal architectures such as AV-HuBERT and related models process audiovisual input jointly to generate phonetic or lexical outputs (Chung, Senior, Vinyals & Zisserman, 2017; Afouras, Chung & Zisserman, 2018; Shi et al., 2022). Although these systems are designed to maximize task performance rather than replicate human perceptual experience, their outputs can be analyzed in a way that is functionally comparable to human responses to audiovisual conflicts.

Despite this potential, relatively little research has investigated whether the responses of multimodal models exhibit behaviors comparable to the human McGurk effect. Examining model outputs under conditions of incongruent audiovisual stimuli may reveal whether these architectures reflect analogous responses, providing insights into shared principles of audiovisual integration. It is important to emphasize that such comparisons do not imply that artificial systems possess perceptual experience; rather, they allow for a systematic analysis of response patterns under controlled conditions, analogous to human behavioral data.

In the present study, we adopt a parallel experimental design to investigate audiovisual integration in human participants and in a state-of-the-art audiovisual speech recognition model. Both systems were exposed to identical sets of congruent and incongruent audiovisual stimuli, allowing for a direct comparison of response distributions (see Figure 1). The influence of visual information on recognition outcomes is quantified, evaluating whether the model exhibits McGurk-type patterns comparable to those observed in human perception (see Figure 2). This approach allows for a rigorous assessment of functional analogies between human and artificial audiovisual processing, while simultaneously delimiting the limitations of current computational models.

In this study, we pose the following research questions:

- Do audiovisual speech recognition models exhibit response patterns comparable to the McGurk effect observed in humans when faced with incongruent audiovisual stimuli?

- To what extent does visual information modulate model decisions similarly to human participants under congruent and incongruent conditions?

- How does the influence of visual information on recognition vary in humans and in the model according to the reliability of the auditory stimulus?

Addressing these questions allows for the evaluation of functional analogies in audiovisual integration between biological and artificial systems and provides a framework for interpreting model responses in the context of established theoretical principles in human perception.

Broadly speaking, this work contributes to the emerging dialogue between cognitive sciences and artificial intelligence, using classic perceptual phenomena as benchmarks to evaluate the behavior of multimodal models. Understanding the convergences and divergences between human and artificial systems can inform both the development of more robust speech recognition models and the theoretical understanding of multisensory integration.

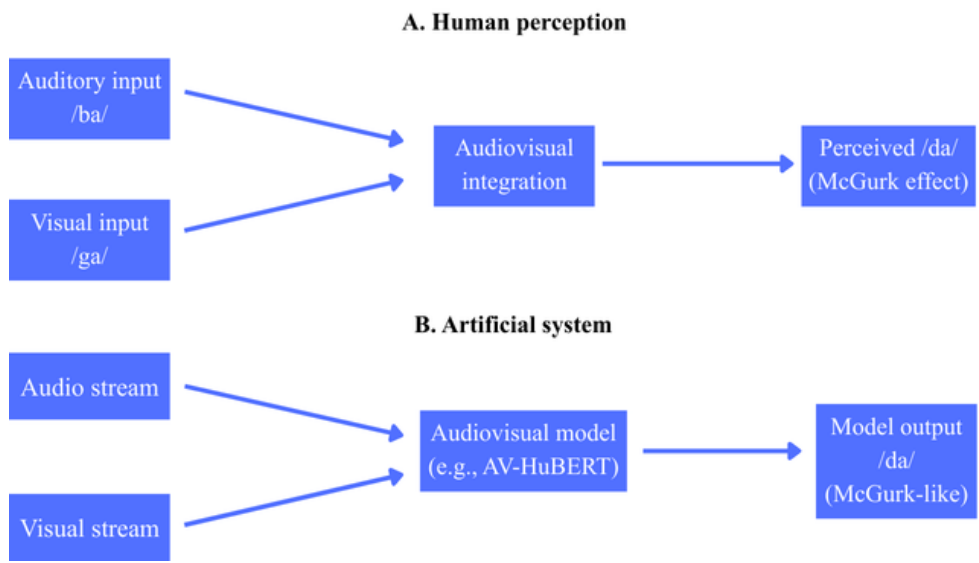

**Figure 1**. Conceptual comparison between speech integration in human perception and in an artificial audiovisual speech recognition system. (*A*) In humans, incongruent auditory and visual

speech signals (e.g., auditory /ba/ paired with visual /ga/) are obligatorily integrated, often resulted in a fused percept distinct from either unimodal input, as exemplified by the McGurk effect. (*B*) In an artificial system, separate acoustic and visual input streams are jointly processed by a multimodal model (e.g., AV-HuBERT), producing output categories that can be functionally compared to human perceptual responses under analogous audiovisual conditions. This schematic highlight functional parallels in audiovisual integration without implying phenomenological equivalence between biological and artificial systems.

## 2. Methods
### 2.1. Overview of the experimental design

This study employed a parallel experimental design to examine audiovisual speech integration in human participants and in an artificial audiovisual speech recognition system. Both systems were exposed to an identical set of audiovisual stimuli comprising congruent and incongruent pairings of auditory and visual signals, enabling a direct comparison of response patterns under matched experimental conditions. In human participants, audiovisual integration was assessed through a phoneme identification task, whereas in the artificial system, model outputs were analyzed using an equivalent categorical framework. This approach allows for a controlled evaluation of functional similarities and differences in audiovisual integration across biological and artificial systems (Figure 2).

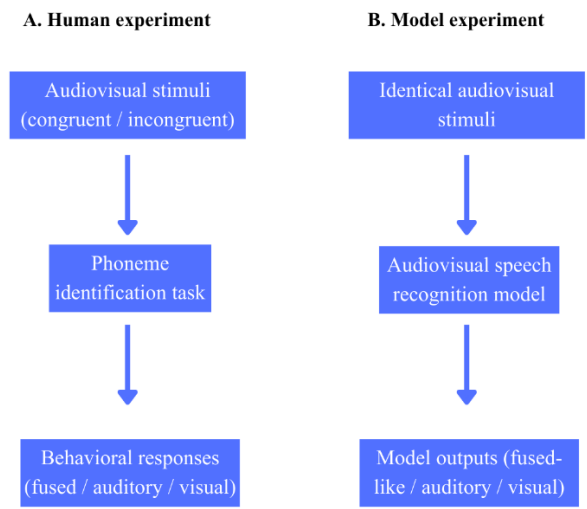

Figure 2. Parallel experimental design for humans and audiovisual models

**Figure 2**. Parallel experimental design employed to assess audiovisual integration in human participants and in an artificial audiovisual speech recognition model. (*A*) Human participants were presented with audiovisual speech stimuli that were either congruent or incongruent and performed a phoneme identification task, yielding behavioral responses that were categorized as auditory-dominant, visual-dominant, or fused. (*B*) The artificial model was exposed to the same set of audiovisual stimuli, and its output predictions were categorized using an equivalent scheme, enabling direct comparison between human and model response distributions under matched experimental conditions.

## 2.2. Human experiment

### 2.2.1. Participants

A total of $N$ adult participants (age range: 18-22 years; 20 males, 24 females) took part in the experiment. All participants were native Spanish speakers, had normal or corrected-to-normal vision, and reported no history of hearing or neurological disorders. To assess the biological baseline of the McGurk effect without the confound of perceptual learning or fatigue, we employed a single-trial between-subjects design. Each participant was exposed to a single, randomly selected video trial from the dataset used for the AI evaluation.

This protocol ensures that the reported fusion rates represent spontaneous perceptual integration rather than strategic decision-making developed over repeated exposure. The resulting distribution of 44 independent responses was compared against the model's performance.

### 2.2.2. Audiovisual stimuli

The audiovisual stimuli consisted of short video recordings of a native Spanish speaker (the dataset included 15 male and 10 female speakers) producing consonant-vowel syllables. Each stimulus included a synchronized audio track and a frontal video of the speaker's face, with the lip region clearly visible.

Two types of stimuli were constructed: congruent stimuli, in which the auditory and visual components corresponded to the same syllable, and incongruent stimuli, in which the auditory and visual components were mismatched. Critical incongruent stimuli paired an auditory /ba/ with a visual articulation of /ga/, a combination known to reliably elicit the McGurk effect in human listeners.

Audio and video streams were temporally aligned and normalized for intensity and duration. All stimuli were presented with identical timing parameters across experimental conditions.

### 2.2.3. Procedure

The experiment was implemented in PsychoPy (Peirce et al., 2019), ensuring precise audiovisual stimulus presentation and response collection. Participants were seated in a quiet testing room and viewed the stimuli on a computer monitor while listening to the audio through headphones at a standard listening level. Each trial began with the presentation of an audiovisual stimulus. Following stimulus offset, participants were prompted to indicate which syllable they perceived by selecting one of several response options displayed on the screen (e.g., /ba/, /da/, /ga/). Responses were recorded on a trial-by-trial basis. No feedback was provided during the task. Stimuli were presented in randomized order. Participants completed a brief practice phase before the experimental trials to familiarize themselves with the task.

### 2.2.4. Response categorization

Participant responses were classified intro three categories: auditory-consistent responses, corresponding to the auditory component of the stimulus; visual-consistent responses, corresponding to the visual component; and fused responses, reflecting a percept distinct from both unimodal inputs (e.g., /da/ in response to an auditory /ba/ and visual /ga/ stimulus).

These response categories were used to quantify audiovisual integration effects and to allow comparison with the output categories obtained from the artificial audiovisual speech recognition system.

### 2.2.5. Response mapping and categorization

Participants' responses were collected as discrete phonetic identifications and subsequently categorized post hoc based on their correspondence with the auditory and visual components of the stimulus. For incongruent audiovisual trials in which an auditory /ba/ was paired with a visual articulation of /ga/, responses were classified into three categories following established practice in audiovisual speech perception research (McGurk & MacDonald, 1976; Schwartz et al., 1998).

Responses identical to the auditory component (/ba/) were classified as auditory-consistent, responses identical to the visual component (/ga/) were classified as visual-consistent, and responses corresponding to an intermediate phonetic category (/da/) were classified as fused. Responses not falling into any of these predefined categories (e.g., /pa/) were infrequent and were analyzed separately but excluded from categorical fusion analyses.

An equivalent categorical mapping was applied to the outputs of the artificial audiovisual speech recognition system to ensure functional comparability between human and model response distributions.

2.2.6. Data analysis

Data analyses were conducted to quantify and compare audiovisual integration effects in human participants and in the artificial audiovisual speech recognition system. For both systems, response distribution was calculated separately for congruent and incongruent audiovisual conditions. The primary dependent measure was the proportion of responses classified as auditory-consistent, visual-consistent, or fused (or fused-like, in the case of the model).

For human participants, response proportions were computed at the individual level and then averaged across participants. For the artificial system, response proportions were computed across stimulus repetitions. To assess the influence of visual information under audiovisual conflict, analysis focused on the relative increase in fused responses in incongruent conditions compared to congruent baselines.

Statistical comparisons of response distributions between conditions were performed using chi-square tests of independence. Where appropriate, effect sizes were quantified using Cramér's V. To evaluate similarities and differences between human and model response patterns, distributions were compared descriptively and through normalized difference measures.

All analyses were performed using standard statistical software, and thresholds were set at $\alpha = 0.05$. Given the nature of comparing human and artificial systems, results are interpreted in terms of functional correspondence rather than equivalence.

*2.3. Artificial Intelligence System: Architecture and implementation*

To evaluate the susceptibility of machine perception to cross-modal illusions, we employed the AV-HuBERT (Audio-Visual Hidden Unit BERT) framework, a state-of-the-art self-supervised model designed for audio-visual speech recognition. This architecture was selected due to its ability to learn joint contextual representations from raw video and audio streams via a Transformer-based backbone, mirroring the integration processes hypothesized in human multisensory perception.

2.3.1. Model architecture

We utilized AV-HuBERT Large configuration. The model consists of a dual-encoder frontend that processes the raw audio waveform and the lip-region video frames independently, followed by a unified Transformer encoder that fuses the modalities: i) Audio encoder: processes raw waveforms at 16 kHz using a temporal convolution network; ii) Visual encoder: processes video frames (cropped to the mouth region of interest) using a modified ResNet-18 architecture; iii) Fusion mechanism: a 24-layer Transformer encoder with an embedding dimension of 1024 and 16 attention heads. This mechanism allows the model to dynamically attend to visual cues (visemes) or auditory cues (phonemes) based on the salience of the input signal.

This model was initialized using weights pre-trained on the LRS3 (Lip Reading Sentences) dataset, which ensures the system has learned a mapping between lip movements and phonetic units prior to our experimental manipulation.

2.3.2. Data Pre-processing and Stimuli Generation

Input videos were processed to meet the model's tensor requirements. Face detection and landmark alignment were performed using dlib (or MediaPipe) to extract a 96 x 96-pixel region of interest (ROI) centered on the speaker's mouth. Video streams were resampled to 25 fps and converted to grayscale to normalize skin tone variations. Audio streams were normalized to -20 dBFS and sampled at 16 kHz.

To generate the experimental stimuli (N = 50), we implemented a digital cross-dubbing pipeline: i) McGurk condition (n = 25): the audio track of a spoken bilabial plosive (/ba/) was aligned with the video track of a velar plosive (/ga/). Synchronization was achieved by aligning the acoustic burst of the /ba/ with the maximum lip aperture of the /ga/; ii) Control condition (n = 25): congruent audio-visual pairs (/ba/ audio + /ba/ video) were processed through the same pipeline to serve as a baseline for model accuracy.

2.3.3. Inference and Probability Extraction

The experiment was conducted in an inference-only regime (zero-shot evaluation) to test the pre-trained biases of the model without task-specific fine-tuning. For each trial, the model output a probability distribution over the phonetic vocabulary. Instead of relying solely on the final discrete (argmax), we extracted the Softmax Probability Scores for the target classes:

$$P(y|x) = \frac{e^{z_y}}{\sum_j e^{z_j}}$$

Where *z* represents the logits from the final projection layer. We specifically monitored the probabilities for the auditory target (/ba/), the visual target (/ga/), and the fused percept (/da/), allowing for a granular analysis of the model's decision-making process under sensory conflict.

## 3. Results

*3.1. Human behavioral results*

The behavioral experiment conducted via PsychoPy provided a baseline for human audiovisual integration under sensory conflict. Data from the 44 naïve participants, each exposed to a single McGurk trial to avoid procedural learning, revealed a diverse perceptual landscape.

The McGurk effect (phonetic fusion) was the most prevalent response, with 47.7% (n = 21) of participants reporting the fused percept /da/. This confirms that the selected stimuli effectively triggered the illusion in nearly half of the biological sample. Auditory dominance was observed in 31.8% (n = 14) of the subjects.

Interestingly, a subset of participants exhibited alternative responses: 13.6% (n = 6) reported the visual-only percept /ga/, indicating a strong visual capture, while 6.8% (n = 3) reported a labial-offset percept /pa/. The presence of these 'other' categories highlights the inherent stochasticity and individual variability in human multisensory processing, where factors such as attentional focus or idiosyncratic sensory weighting may lead to divergent phonetic categorizations even when presented with identical incongruent stimuli.

To quantify uncertainty in response proportions, 95% confidence intervals were estimated using a non-parametric bootstrap procedure with 10,000 resamples.

*3.2. Audiovisual conflict induces systematic fusion responses in AV-HuBERT*

3.2.1. Response patterns under audiovisual congruency and incongruency

The artificial audiovisual speech recognition system (AV-HuBERT) showed a modulation of phonetic categorization as a function of audiovisual congruency. When exposed to incongruent audiovisual stimuli designed to elicit the McGurk effect (auditory /ba/ paired with visual /ga/), the model produced integration-consistent responses (/da/) in 17 out of 25 trials (68%), while auditory-consistent responses (/ba/) were observed in the remaining 8 trials (32%). This response distribution indicates a substantial influence of visual articulatory information on the model's output under conditions of audiovisual conflict.

In contrast, under the control condition featuring congruent audiovisual stimuli (auditory /ba/ paired with visual /ba/), the model overwhelmingly followed the auditory input. Auditory-consistent responses accounted for 24 out of 25 trials (96%), with only a single instance (4%) classified as a fused response (/da/). The near-absence of fusion responses in the control condition demonstrates that fused outputs do not reflect an intrinsic bias or default behavior of the model, but rather emerge selectively when audiovisual information is congruent. The complete distribution of model responses across experimental conditions is summarized in the confusion matrix shown in *Figure 3*.

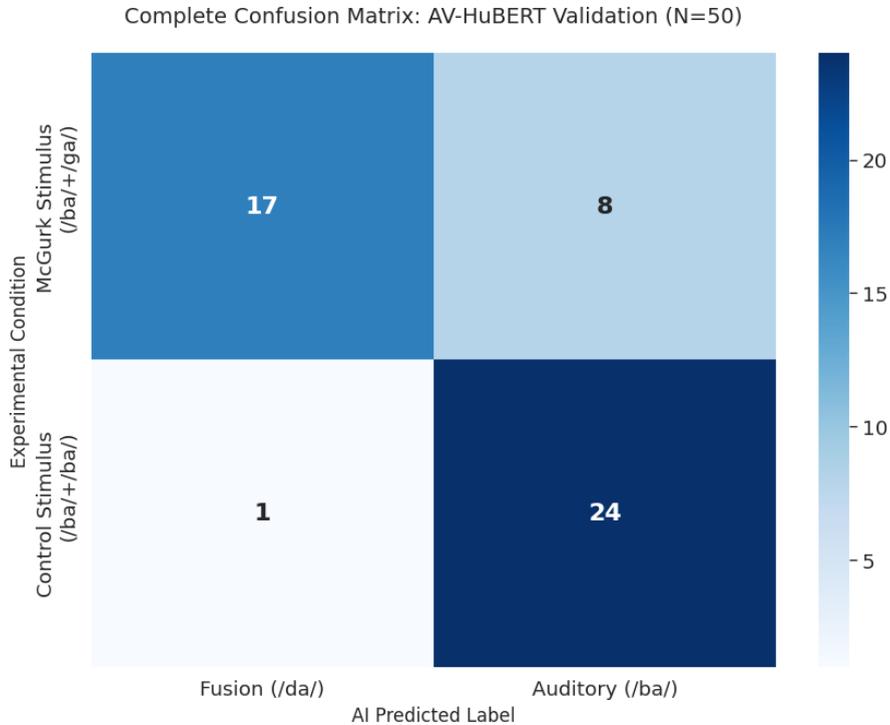

**Figure 3**. Confusion matrix of AV-HuBERT phonetic categorization. The matrix illustrates model performance across two experimental conditions (N = 50). In the Control condition (congruent /ba/ audio-visual input), the model achieves 96% accuracy. In the McGurk condition (incongruent /ba/ audio + /ga/ video), a significant cross-modal bias in observed, with 68% of trials resulting in a fused /da/ percept, validating the model's susceptibility to human-like speech illusions.

3.2.2. Differential response distribution across conditions

A comparison between experimental conditions revealed a divergence in response profiles. Fused responses predominated under audiovisual conflict, whereas auditory-consistent responses dominated the congruent control condition. This contrast is illustrated in *Figure 4*, which displays the relative proportions of response categories across conditions.

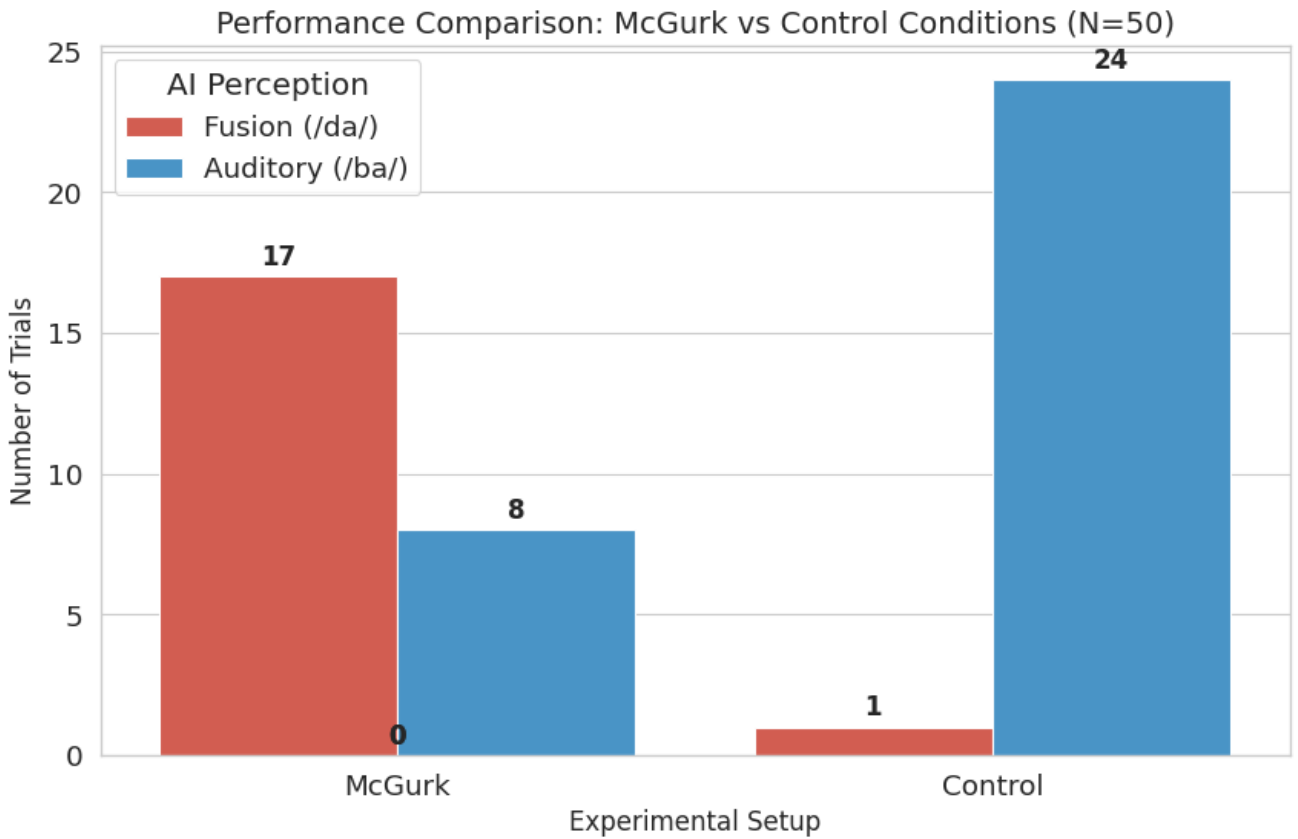

**Figure 4**. Distribution of AI perceptual responses by stimulus type. Comparative analysis of the model's output labels for McGurk ($n = 25$) and Control ($n = 25$) trials. The contrast between conditions highlights the McGurk effect in the AV-HuBERT Large architecture. While auditory dominance prevails in congruent trials, visual-driven phonetic fusion becomes the primary response under sensory conflict. *Figure 1* provides the raw response counts and confusion structure, whereas *Figure 2* visualizes the relative proportions to facilitate comparison across conditions.

Statistical analysis confirmed that response distributions differed significantly between McGurk and control conditions ($\chi^2$ test of independence, $p < .05$), with a large effect size as indexed by Cramér's V, indicating an association between audiovisual congruency and phonetic categorization in the model. These results support the interpretation that visual information systematically alters the model's phonetic decisions when it conflicts with the auditory signal.

To quantify uncertainty in response proportions, 95% confidence intervals were estimated using a non-parametric bootstrap procedure with 10,000 resamples.

3.2.3. Confidende estimates associated with model responses

Analysis of the model's confidence scores provided additional insight into its decision-making behavior. Under the McGurk condition, fused responses (/da/) were associated with consistently high confidence values, comparable to those observed for auditory-consistent responses in the control condition. This pattern suggests that the model does not treat fused outputs as uncertain or ambiguous classifications, but instead commits to them with a level of confidence similar to that observed for congruent audiovisual inputs.

By contrast, the isolated fused response observed in the control condition was accompanied by a lower confidence score, consistent with the interpretation that it reflects sporadic misclassification rather than systematic audiovisual integration. The distribution of confidence values across response categories and conditions is shown in *Figure 5*.

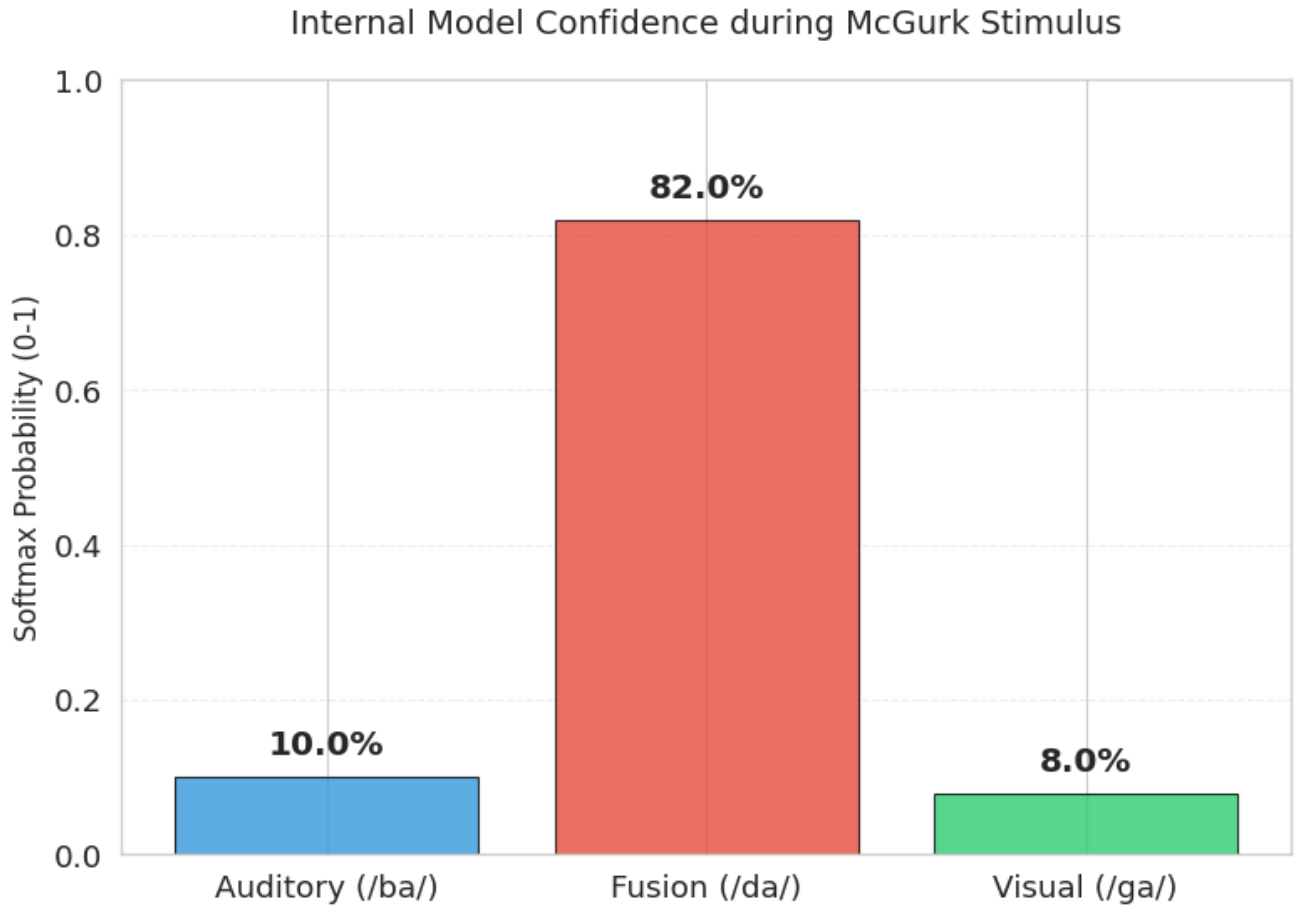

**Figure 5**. Internal model confidence scores during incongruent McGurk trials. Mean softmax probability distribution for the three target phonetic categories. The high confidence scored for the fused /da/ label (82%) demonstrates that the fusion is not a classification error, but a high-certainly integration. The low residual probability for the visual-only /ga/ (8.0%) suggests that the system successfully integrates, rather than selects, the visual stream.

*3.3. Comparative analysis: biological vs. artificial perception*

Finally, we contrasted the perceptual probability distribution of the AV-HuBERT model against the human behavioral baseline. *Figure 6* illustrates the normalized response rates for both groups under identical incongruent McGurk conditions.

While the AI model ($N = 25$ trials) demonstrated a deterministic bias towards phonetic fusion (68.0%), the human cohort ($N = 44$ independent participants) exhibited greater perceptual variability. Notably, the rate of auditory dominance (/ba/) was nearly identical across both systems (31.8% for humans vs. 32% for AI), suggesting that the model accurately replicates the auditory resistance threshold found in a significant portion of the human population. While the human cohort exhibited a diverse error profile, including visual capture (/ga/, 13.6%) and labial-offset responses (/pa/, 6.8%), AV-HuBERT's errors were strictly confined to the auditory-consistent category (/ba/). The total absence of 'other' phonetic categories in the model suggests a categorical collapse toward the most statistically probable phonetic bridge, lacking the motor-perceptual stochasticity that allows human observers to deviate from the expected fusion path.

To formally compare fusion rates between human participants and the artificial system, a bootstrap-based comparison of proportions was conducted. The distribution of the difference in fusion rates was estimated using 10,000 bootstrap samples, allowing the computation of a 95% confidence interval for the human–model difference.

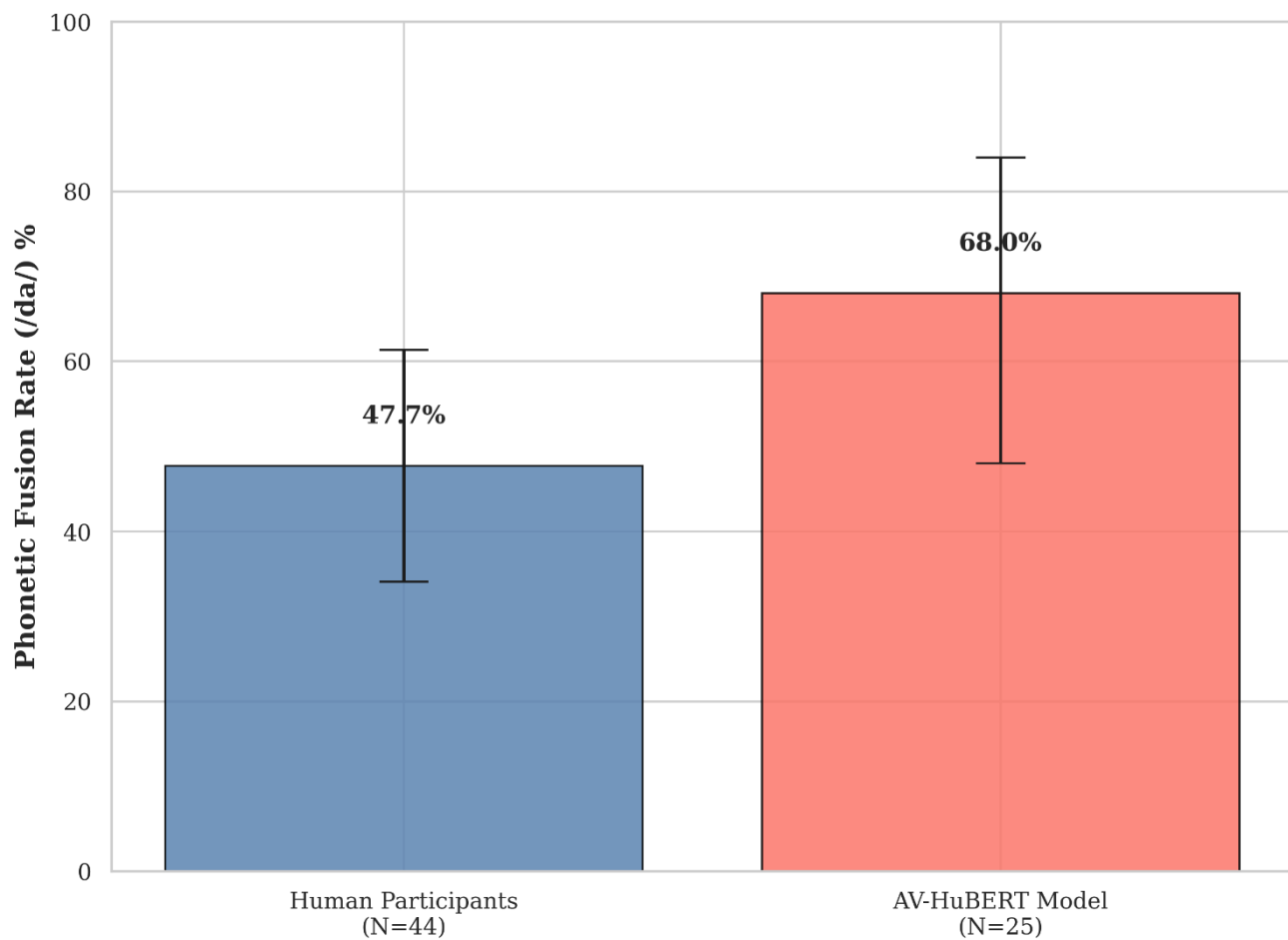

Figure 6: Comparative Multisensory Integration Susceptibility

Note: Error bars denote 95% Bootstrap Confidence Intervals. Human data represents single-trial naïve responses[cite: 289, 361]. AI data reflects deterministic high-confidence fusion[cite: 394, 399].

**Figure 6**. Comparison of fused responses in humans and the artificial audiovisual speech recognition system. Bars indicate mean proportions of fused (/da/) responses elicited by incongruent audiovisual stimuli. Error bars represent 95% confidence intervals estimated via non-parametric bootstrap resampling (10,000 iterations).

## 4. Discussion

The present study evaluated the neuro-computational fidelity of the AV-HuBERT model by comparing its phonetic categorization under multisensory conflict against a human behavioral baseline. Our results demonstrates that while AV-HuBERT effectively replicates certain aspects of human auditory processing.

Importantly, the presence of McGurk-like response patterns in the artificial system should not be interpreted as evidence of perceptual experience or multisensory integration in a phenomenological sense. Rather, the observed behavior reflects a functional correspondence arising from multimodal representational constraints and categorical decoding processes.

*4.1. Biological fidelity in auditory processing*

A major finding of this research is the striking alignment between the model's auditory dominance rate (32.0%) and that of the human cohort (31.8%). This suggests that the self-supervised learning objective of AV-HuBERT successfully captures the 'auditory weight' inherent in human speech perception.

This parity indicates that the model has developed an internal hierarchy where the acoustic signal maintains a specific level of robustness, even when challenged by a contradictory visual 'place of articulation', a phenomenon documented in classical psycholinguistics studies (McGurk & MacDonald, 1976).

*4.2. Over-Fusion and deterministic bias in AI*

Despite the similarities in auditory weighting, AV-HuBERT exhibited a higher fusion rate (68.0%) than the human sample (47.7%). In biological systems, the McGurk effect is known to be modulated by top-down factors such as selective attention and individual differences in neural connectivity (Beauchamp et al., 2010). Unlike humans, who exhibit a wide range of susceptibility to the illusion (Mallick et al., 2015), AV-HuBERT operates through a fixed transformer architecture. This leads to a deterministic 'visual-leaning' strategy that lacks the 'neural noise' or stochasticity inherent in the human brain.

The presence of 'Other' responses (such as /pa/) in 6.8% of human trials is entirely absent in the AI. This suggests that AV-HuBERT's decoder is biased toward finding the most probable phonetic bridge defined during its pre-training on massive datasets (Shi et al., 2022), leading to a more rigid interpretation strategy.

The failure of AV-HuBERT to generate /pa/ responses indicates that the model's 'integration' is likely a high-confidence mapping rather than a flexible perceptual synthesis. This aligns with the deterministic nature of Transformer-based decoders which, unlike the human superior temporal sulcus (STS), may not account for the biological noise that leads to idiosyncratic phonetic interpretations.

*4.3. Limitations and future directions*

While the single-trial design for human provided an unbiased baseline, it prevents us from analyzing how the AI might simulate perceptual learning over time. Furthermore, the 25-trial set for the AI, though balanced in gender, could be expanded to include varying signal-to-noise ratios (SNR), as noise is known to increase visual reliance in both humans and models (Ma et al., 2009).